\documentclass[11pt,a4paper]{article}
\usepackage[hyperref]{acl2021}
\usepackage{times}
\usepackage{latexsym}
\usepackage{tabularx}
\usepackage{booktabs}
\usepackage{arydshln}  
\usepackage{multirow}
\usepackage{makecell}

\usepackage{microtype}

\aclfinalcopy 

\begin{document}

\title{Evaluating CxG Generalisation in LLMs\\ via Construction-Based NLI Fine Tuning}

\author{%
\textbf{Tom Mackintosh$^1$,
Harish Tayyar Madabushi$^1$,
Claire Bonial$^2$,} 
\\
$^1$University of Bath, U.K. \\
$^2$DEVCOM Army Research Laboratory, U.S.A. \\
\texttt{htm43@bath.ac.uk}
}

\maketitle
\begin{abstract}
We probe large language models' ability to learn deep form-meaning mappings as defined by 
construction grammars. We introduce the ConTest-NLI benchmark of 80k sentences covering eight English constructions from highly lexicalized to highly schematic. Our pipeline generates diverse synthetic NLI triples 
via templating and the application of a model-in-the-loop filter. This provides aspects of human validation to ensure challenge and label reliability. Zero-shot tests on leading LLMs reveal a 24\% drop in accuracy between naturalistic (88\%) and adversarial data (64\%), with schematic patterns proving hardest. Fine-tuning on a subset of ConTest-NLI yields up to 9\% improvement, yet our results highlight persistent abstraction gaps in current LLMs and offer a scalable framework for evaluating construction-informed learning.
\end{abstract}

\section{Introduction and Motivation}

\begin{table}[h!]
\centering
\setlength{\tabcolsep}{4pt} 
\begin{tabular*}{\columnwidth}{@{\extracolsep{\fill}}lcc}
\toprule
\textbf{Model} & \textbf{\makecell{Constr.\\Semantics}} & \textbf{\makecell{Constr.\\Distinction}} \\
\hdashline
\multicolumn{3}{c}{Prior work~\citep{scivetti2025assessing}}\\
\hdashline
GPT-4o      & 0.88 & 0.58 \\
GPT-o1      & 0.90 & 0.46 \\
Llama 3 70B & 0.74 & 0.52 \\
\hdashline
Human       & 0.90 & 0.83 \\
\midrule
\multicolumn{3}{c}{This work}\\
\hdashline
\makecell{Llama-3.1-8B \\ (baseline)} & 0.57 & 0.33 \\
\makecell{Llama-3.1-8B \\ (fine-tuned)} & 0.66 & 0.39 \\
\bottomrule
\end{tabular*}
\caption{Comparison of model performance on constructional (constr.) understanding. The top section, with results from prior work~\citet{scivetti2025assessing}, shows that LLMs struggle with the constructional distinction task compared to the human baseline. The bottom section presents our results, showing that this shortcoming persists despite fine-tuning. See Section \ref{sec:results} for full results.}
\label{tab:results_single_col}
\end{table}

Human intelligence is often attributed to our capacity for language — and, in particular, our ability to generalize abstract, compositional meaning from surface structure~\cite{10.1093/acprof:oso/9780199244843.003.0002}. Construction Grammar (CxG)~\citep{goldberg1995constructions, croft2001radical, madabushi2020cxgbert} (See also Section \ref{sec:relwork}) formalises this by treating linguistic knowledge as form-meaning pairings — constructions — that range from single words to complex syntactic frames. Understanding whether large language models (LLMs) acquire such abstractions remains a fundamental question at the intersection of linguistics and artificial intelligence.

In CxG, each construction pairs a conventionalised \emph{form} with an associated meaning. The form is the syntactic configuration, possibly including fixed lexical items, while the meaning is provided by the construction as a whole rather than from the individual lexical items. For example, the Resultative construction has the form Noun Phrase (NP), Verb (V), Noun Phrase (NP), Adjective (ADJ) and the meaning ``the action described by the verb causes the object to enter the state described by the adjective'' \cite{goldberg1992argument}. In ``She hammered the metal flat,'' the state `flat' is the \emph{result} of the hammering event, a meaning supplied by the Resultative. 

While each construction has a specific \emph{form}, different constructions can share the same syntactic structure. For instance, the Depictive construction also uses the NP V NP ADJ \emph{form} but has a distinct \emph{meaning} \cite{goldberg2004english}. In the Depictive, the adjective describes the state of the noun \emph{during} the action of the verb, not as a result of it. This is illustrated by the example, ``A famous emperor buried scholars alive.'' Here, `alive' describes the state of the scholars while they were being buried; crucially, the act of burying did not \emph{cause} them to become alive. This distinction highlights how syntactically identical sentences can convey vastly different meanings based on the underlying construction.

\begin{table}[t]
\centering
\setlength{\tabcolsep}{3pt}
\footnotesize
\begin{tabularx}{\columnwidth}{lXXc}
\toprule
\textbf{Construction} & \textbf{Premise} & \textbf{Hypothesis} & \textbf{Label} \\
\midrule
Resultative &
Through effort, the gardener cultivated the garden lush. &
The gardener worked hard to create a vibrant outdoor space. &
Entailment \\
Caused Motion &
The magician levitated the rabbit into the hat. &
The magician placed the rabbit on the table. &
Contradiction \\
Causative With &
In no time, the magician had filled the auditorium with applause. &
The magician performed in different auditoriums. &
Neutral \\
\bottomrule
\end{tabularx}
\caption{Examples drawn from the ConTest-NLI training set, with one instance of each NLI label from distinct constructions.}
\label{tab:cxg_nli_intro_examples}
\end{table}

Recent evaluation work~\citep{scivetti2025assessing}, which used the downstream task of Natural Language Inference (NLI) to create a test of the functional understanding of LLMs, reveals that while LLMs can correctly interpret an entrenched construction like the Resultative  even with unusual lexical items, their generalization ability is limited. Specifically, when presented with creative instances of a less entrenched construction like the Depictive, LLMs tend to overgeneralize and assign the meaning of the more frequent, or entrenched, Resultative construction. Indeed they overgeneralise to such an extent that they show a performance drop of over 40\% on on this task, when compared to the original task of interpreting the meaning of entrenched constructions. These findings are summarized in Table~\ref{tab:results_single_col}.  This failure to use lexical and pragmatic cues to resolve syntactic ambiguity, a task at which native speakers can perform quite easily, demonstrates that the models' grasp of abstract meaning remains brittle and overly dependent on statistical patterns rather than a robust, human-like linguistic competence.

While \citet{scivetti2025assessing} identify this shortcoming, they leave the reasons for this specific failure to future work. Therefore, this work aims to answer this question by examining the model's expertise with the task. Specifically, \textbf{we hypothesize and investigate if training the model on explicit NLI examples will help the model better `understand' creative, less entrenched constructions in the presence of a more frequent distractor.} A positive result would offer a clear-cut path to improving these models' understanding, whereas a negative result would point to a more fundamental issue that needs to be addressed.

To this end, this paper introduces \textbf{ConTest-NLI} (Constructional Test Natural Language Inference), a scalable dataset designed to evaluate whether LLMs internalize the semantics of linguistic constructions or rely solely on surface heuristics. ConTest-NLI specifically targets systematic generalization across unseen verbs, arguments, and constructions. This provides a scalable way to inform LLMs of specific construction examples, allowing a new control for deeper research into semantic understanding of linguistic theory.

One key empirical finding is that \textbf{LLMs fail to generalize constructional semantics across syntactically identical but semantically distinct constructions}. For example, models trained to detect entailment violations in the Resultative construction show no improved performance on the Depictive construction, despite their shared syntax. This lack of transfer reveals that current models do not acquire construction-general semantics, but instead overfit to narrow instantiations.

To test our hypothesis at scale, we use a semi-automated pipeline that facilitates generation of synthetic constructional NLI triples: ConTest-NLI. Example data is shown in Table~\ref{tab:cxg_nli_intro_examples}.  Our pipeline leverages syntax-informed template generation of eight core constructions and model-in-the-loop filtering to identify deceptive false positives.

We compare ConTest-NLI to two existing CxG benchmarks from \citet{scivetti2025assessing}: the manually curated Construction-NLI (CxN-NLI), and the more challenging Construction-NLI-Distinction (CxN-NLI-Distinction), which introduces false positives that share syntax but diverge in semantics. While those datasets offer excellent linguistic control, they remain small and difficult to scale. ConTest-NLI complements them by enabling controlled experiments across a broader constructional space, yielding more robust insights into model generalization.

We fine-tune small-scale LLMs (LLaMA 3.1 8B Instruct, Mistral 8B Instruct) on ConTest-NLI examples and evaluate their performance across both seen and unseen constructions. While models improve ($\leq$9pp) on the trained construction, their failure to generalise — especially to constructions with shared syntactic structure — suggests a fundamental limitation in semantic abstraction.

\textbf{ConTest-NLI} is thus shown to be useful for evaluating systematic language understanding in LLMs, bridging the scale of automated generation with the precision of theoretical linguistics. In our experiment, we use ConTest-NLI to gather direct empirical evidence that shows, without further architectural or training innovations, LLMs do not acquire transferable constructional semantics --- highlighting a key divergence from human-like generalization.

\section{Related Work}
\label{sec:relwork}

CxG is a linguistic theory that positions constructions --- form-meaning pairings --- as 
the fundamental units of language. A construction, as defined within this framework, is any 
linguistic pattern whose meaning is not fully predictable from its individual components 
\citep{goldberg1995constructions, croft2001radical, madabushi2020cxgbert}.

Further cognitive and usage-based studies within CxG emphasize that humans generalize constructional 
meanings from frequency of exposure and exemplar experiences. Psycholinguistic research, notably by 
\citet{bencini2000contribution}, showed that participants' interpretations of sentence meanings 
significantly reflect constructional semantics rather than just verb meanings alone. In their 
experiment, participants grouped sentences primarily by the underlying constructional meaning, 
demonstrating that constructions themselves carry cognitive reality independent of specific lexical 
content \citep{kaschak2007long, goldberg2007constructions}.

This perspective is particularly relevant for evaluating language comprehension in computational 
models. Recent computational linguistic research leverages CxG to systematically assess language 
understanding in large language models. Studies such as those by \citet{madabushi2020cxgbert} and 
\citet{scivetti2025assessing} illustrate how CxG provides a robust theoretical grounding to create 
targeted, semantically-rich evaluations for LLMs. These studies specifically demonstrate the 
utility of construction-based Natural Language Inference (NLI) tasks, highlighting significant 
limitations of LLMs in generalizing abstract constructional semantics when faced with novel 
linguistic contexts or minimally represented constructions \citep{bonial2024constructing}.

Thus, CxG not only provides insights into human linguistic competence but also offers a rigorous 
toolset for probing and understanding the boundaries of true semantic generalization in language 
models --- a foundational concern of contemporary NLP research.

Also, constructional semantics provide a controlled yet diverse linguistic testbed. Constructions 
vary significantly in their schematicity --- from highly substantive, lexically fixed forms, such as 
the Let-alone construction, to more abstract and schematic patterns such as Resultative or Caused-motion 
constructions \citep{bonial2024constructing, scivetti2025assessing}. Evaluations across this 
spectrum enable systematic testing of LLMs' capacity for abstract semantic generalization. 
Crucially, previous computational studies demonstrate that while LLMs may perform well on lexically 
anchored constructions due to frequency and memorization, their performance substantially 
deteriorates when faced with more schematic and less frequent constructions 
\citep{weissweiler2022better, scivetti2025assessing}.

Despite the promise of construction grammars as a diagnostic for true semantic generalization, 
existing computational CxG evaluations remain narrowly focused, limited in scale, or insufficiently 
controlled. This project fills that gap by introducing a semi-automated pipeline that combines high-variance templating to produce large-scale CxG evaluation data. 


\section{ConTest-NLI Dataset Development}

ConTest-NLI is designed as a scalable, high-variance training resource for probing whether LLMs can learn and generalise the semantics of English constructions beyond rote lexical recall. It forms the centrepiece of a broader multi-corpus strategy, enabling both in-domain fine-tuning and rigorous, out-of-distribution testing. This is essential: single-source datasets are prone to heuristic exploitation, whereas orthogonal axes --- synthetic vs.\ natural, fluent vs.\ adversarial --- allow us to pinpoint exactly where generalisation succeeds or fails.

We adopt the eight English constructions from \citet{scivetti2025assessing}, spanning the substantive--schematic continuum (e.g.,\ Let-alone vs.\ Resultative). Construction details and examples are provided in Appendix~\ref{appendix:constructions}.  Each construction is instantiated by $\geq$10{,}000 examples, generated from $\geq$8 canonical templates varying surface order, clause type, and optional modifiers. 

\subsection{Template Engineering}

For each construction, we hand-crafted $8$--$12$ canonical skeletal templates encoding the obligatory syntactic positions and any construction-specific function words (e.g.,\ ``The more X, the more Y'' for the Comparative Correlative). These templates are designed to maximise \emph{controlled diversity}: varying word order, clause type, voice (active/passive), adjunct position, and optional modifiers ensures that no single surface pattern dominates. 

Lexical slots are populated from “mid-frequency” lemmas (20--60th percentile in Book\-Corpus) to reduce overlap with model pretraining data. We further expand these lists using WordNet synonyms, hyponyms, and antonyms, while explicitly excluding the top 10{,}000 Common Crawl tokens and any lexemes whose semantics would trivially satisfy the inference (e.g.,\ \emph{moved} in a Caused-Motion frame). Controlled adverbial pools (manner, time, frequency, intensity) and automated morphological inflection via \texttt{lemminflect} add stylistic variation without altering truth-conditional content. Examples of templates and their instantiations are provided in Table~\ref{tab:contestnli-examples}, and templates for all constructions are provided in Appendix~\ref{appendix:templates}.

\begin{table}[ht]
\centering
\small
\begin{tabular}{p{1.6cm}p{5.2cm}}
\toprule
\textbf{Construction} & \textbf{Example Template / Instantiation} \\
\midrule
\textsc{Resultative} & ``The [agent] [verb] the [patient] [end-state]'' $\rightarrow$ \emph{The chef chopped the carrots thin} \\
\textsc{Caused-Motion} & ``X [verb] Y into Z'' $\rightarrow$ \emph{They rolled the log into the river} \\
\textsc{Causative-With} & ``X filled C with S'' $\rightarrow$ \emph{The artist filled the gallery with vibrant paintings} \\
\textsc{Let-Alone} & ``Even getting X to [verb] was tough, let alone Y'' $\rightarrow$ \emph{Even getting the robot to succeed was tough, let alone the knapsack} \\
\bottomrule
\end{tabular}
\caption{Sample templates and instantiations from the ConTest-NLI generation pipeline. Note that template filling results in some semantic infelicity, such as the Let-alone comparison of a robot and a knapsack. }
\label{tab:contestnli-examples}
\end{table}

\subsection{Example Generation}
Each premise sentence is paired with three hypotheses labelled \emph{entailment} (E), \emph{neutral} (N), or \emph{contradiction} (C), with labels assigned via construction-specific generation rules grounded in formal semantics. For example, a \textsc{Causative-With} premise \emph{The artist filled the gallery with vibrant paintings} yields:
\begin{itemize}
    \item (E) \emph{The gallery contained vibrant paintings}
    \item (C) \emph{The gallery was empty of any paintings}
    \item (N) \emph{The artist painted in a nearby studio}
\end{itemize}
This approach ensures that all examples are fluent and natural-sounding, while still requiring the model to attend to the construction’s form-meaning pairing to make the correct inference.

\subsection{Manual Analysis}
We conducted manual analysis to ensure the quality of the dataset along two dimensions: (i) the generated constructions are indeed members of the specified construction type, and (ii) the established relation for each NLI triple is accurate. We randomly sampled 100 instances of our dataset, balanced across neutral, entailment, and contradiction relations. One author and native English speaker, trained in linguistics and CxG, provided a binary rating for (i) and (ii), and where the author disagreed with the relation provided, gave a corrected NLI relation.  The result of this analysis was that 99/100 instances were judged to be instances of the specified construction type, and 94/100 NLI instances were judged to have the correct relation.  This indicates the overall high quality of the developed dataset. 

However, the manual analysis further revealed two limitations of the synthetic NLI triples. First, the data in the sample were relatively repetitive. While we expect repetitions of the premise with unique hypotheses representing different entailment relations to the premise, we found that the hypotheses themselves were also somewhat repetitive, sometimes differing only in a single word (e.g., ``tree trunk'' vs. ''tree bark'' or adding ``might'').  Second, judging the entailment relation was somewhat trivial for many triples, given that entailed hypotheses were sometimes near-verbatim repetitions of the premise, whereas contradicted hypotheses often leveraged a single lexical item of opposite semantics to a counterpart in the premise. We note that manual development of NLI triples can also lead to the same limitations.  

\subsection{Dataset Splits}
We enforce a deterministic $70$/$15$/$15$ train/dev/test split \emph{within each construction}. Crucially, the split is \emph{lexeme-held-out}: any verb lemma appearing in the test set for a given construction is entirely absent from its train and dev sets. This protocol is applied consistently across all ConTest-NLI variants and related evaluation sets (CxN-NLI, CxN-NLI-Distinction), ensuring that improvements can be attributed to constructional abstraction rather than memorisation of specific lexical fillers.

Each construction is balanced across the three NLI labels, yielding 4{,}000 triples per construction and a total of 32{,}000 examples. The class balance ensures that macro-accuracy remains an unbiased measure of model performance.

\section{Fine-Tuning Method}
We use a small variety of base models for our fine-tuning experiments: Llama-3.1-8B-Instruct
and Mistral-8B-Instruct. These models, both with approximately 8.1 billion parameters, have a decoder-only transformer architecture that has already undergone instruction tuning, making them proficient at 
following natural language prompts. Their relatively modest size allows for experimentation 
on single GPU setups, while their strong zero-shot baseline performance on tasks like NLI ensures that any observed gains from fine-tuning are both conservative and meaningful. We also experiment with GPT-4o, however, due to the closed nature of the model, and the high costs of fine-tuning, we test GPT-4o in the 3-shot ICL setting, as an approximation for fine-tuning. We provide hyperparameters in Appendix~\ref{appendix:hyperparameters}, full fine-tuning details in Appendix~\ref{appendix:fine-tuning}, and training regimes in Appendix~\ref{appendix:training-regimes}.

\section{Evaluation Framework}
To rigorously assess our hypothesis that targeted fine-tuning yields systematic constructional understanding a comprehensive evaluation framework is employed. This framework specifies the core metrics for NLI tasks, outlines the use of diagnostic benchmarks to guard against overfitting and ensure generalization, and details essential controls and sanity checks to validate the genuineness of observed performance gains. Our primary metric to measure success is macro-accuracy. A statistically significant improvement of over 5\% accuracy over a model's baseline evaluation (before fine-tuning)
would be sufficient for us to accept our hypothesis. 

\subsection{Diagnostic Benchmarks}
To ensure that improvements are not merely task-specific overfitting but represent genuine, 
transferable gains in understanding, performance on diagnostic benchmarks is critical.
For this purpose we use \citet{scivetti2025assessing} the previously described CxN-NLI and CxN-NLI-Distinction datasets. These benchmarks  feature out-of-distribution compositional tasks that
involve the eight constructions targeted in fine-tuning; however, they are hand-crafted to
test semantic understanding of the constructions.

If a model exhibits consistent performance across all of our datasets, yet remains consistent
in these diagnostic benchmarks, we can confidently claim the model has improved on
constructional usage; however, has not improved on the true understanding of the construction.

\section{Results and Discussion}
\label{sec:results}
\begin{table}[t]
    \centering
    \scriptsize
    \setlength{\tabcolsep}{4pt}
    \begin{tabular}{lccccc}
        \toprule
        \textbf{Model} & \textbf{Setting} & \textbf{CxN-NLI} & \textbf{CxN-NLI-Distinction} \\
        \midrule
        \multirow{2}{*}{LLAMA-3.1-8B} & baseline & 57 & 33 \\
        & fine-tuned & 66 & 39 \\
        \midrule
        \multirow{2}{*}{Mistral-8B} & baseline & 49 & 36 \\
        & fine-tuned & 63 & 37 \\
        \midrule
        \multirow{2}{*}{GPT-4o} & baseline & 88 & 64 \\
        & 3-shot ICL & 91 & 65 \\
        \bottomrule
    \end{tabular}
    \caption{Results across CxN-NLI and CxN-NLI-Distinction benchmarks using baseline and ConTest-NLI fine-tuned models or in-context-learning (ICL) examples from ConTest-NLI.}
    \label{tab:results}
\end{table}

ConTest-NLI demonstrated systematic gains across the CxN-NLI evaluation set; however,
showed only minimal, unsystematic improvements at semantic understanding of the CxN-NLI-Distinction dataset. Results are summarized in Table~\ref{tab:results}. 

Ultimately, this shows model reasoning is done on surface-cues of constructions, rather than true constructional understanding. Critically, we know this is fundamentally different from human reasoning, where we are able to grasp the semantics of constructions instead of just surface-cues.

Notably, all models we test, when fine-tuned on ConTest-NLI, showed a notable increase in accuracy on the CxN-NLI dataset. LLaMA-3.1-8B showed an increase from 57\% to 66\% and Mistral-8B-Instruct 
from 49\% to 63\%. These improvements comfortably exceeded the hypothesized +5 percentage 
point threshold. Similarly, GPT-4o shows an improvement from 88\% to 91\% when 3 in-context examples are added to the prompt. 

Simultaneously, however, the performance increase on the CxN-NLI-Distinction dataset, as a consequence of fine-tuning. is consistently lacklustre. Mistral-8B gains only 1pp after fine-tuning and LLAMA-3.1-8B improves from 33\% to 39\%, a six percentage point increase. While the improvement of LLAMA-3.1-8B might appear to be not insignificant, it should be noted that this improvement is from 33\% which is the random baseline. Similarly, GPT-4o shows an improvement of only 1pp when 3 in-context examples are added to the prompt. More importantly, despite fine-tuning, our results demonstrate a significant drop in performance between the CxN-NLI dataset and the CxN-NLI-Distinction, unlike humans who show only a small performance drop.

\subsection{Error Analysis}
\label{ssec:error-analysis}
We provide full error analysis in Appendix~\ref{appendix:error-analysis}.  The six examples outlined in Table \ref{tab:errors}, each drawn from a different construction in the ConTest-NLI training set, illustrate the central weakness our paper identifies: the model’s reliance on surface-level lexical and syntactic cues rather than robust, construction-general semantic reasoning. In each case, the model either (a) overfit to familiar lexical frames without integrating their semantic consequences, (b) failed to connect constructional form to the entailments it licenses (e.g., \textit{way}-manner implying location change, resultatives implying caused state), or (c) ignored clear scalar or negation cues when they appeared in less frequent or slightly varied contexts. That these errors occur across all eight constructions --- rather than being isolated to a single form --- reinforces our quantitative finding: fine-tuning improved in-domain recognition but did not instill transferable, abstract constructional understanding.

\subsection{Summary and Discussion}
Our ConTest-NLI results illustrate that fine-tuning on natural-sounding premises yields in-domain accuracy gains --- key evidence that LLMs can internalize form-meaning mappings and structures when they encounter sufficiently varied, human-plausible NLI examples. Despite this, our results demonstrate that  their generalization ability is limited. Specifically, when presented with creative instances of a less entrenched construction (e.g., Depictive) LLMs tend to overgeneralize and assign the meaning of the more frequent, or entrenched construction. 

\section{Future Work and Conclusions}

Our investigation demonstrates that explicitly grounding LLM supervision in 
CxG yields measurable gains in systematic generalization, yet also exposes persistent limits of current models' abstraction capabilities. By fine-tuning small-scale LLMs on a CxG-informed corpus --- ConTest-NLI --- we show that targeted constructional supervision delivers substantial improvements (9\% on existing CxG-NLI benchmarks), but that these gains attenuate on out-of-distribution and adversarial challenge items (CxN-NLI-Distinction dataset). These findings carry two broader implications for cognitive modeling.

First, our results suggest that, unlike human learners who extract and re-apply abstract 
form-meaning pairings across lexemes and structures, LLMs continue to rely on residual 
surface cues even after targeted fine-tuning. This divergence highlights the need for 
cognitive models of learning to account for both exemplar-driven acquisition and the 
development of schematic templates, offering a new benchmark against which to evaluate 
theories of human grammatical abstraction.

Second, the semi-automated pipeline we introduce --- combining 
model-in-the-loop adversarial filtering and human validation --- provides a scalable 
methodology for instilling constructional knowledge in models. Integrating such 
CxG-grounded datasets into training regimens can drive more robust semantic 
generalization, informing future architectures that more closely mirror human-like 
compositional reasoning.

\bibliographystyle{acl_natbib}
\bibliography{anthology,acl2021}

\appendix

\section{Constructions of Focus}
\label{appendix:constructions}
The constructions that we develop training data and test on are detailed in Table~\ref{tab:construction-table}.
\begin{table*}[t]
    \centering
    \renewcommand{\arraystretch}{1.2} 
    \begin{tabular}{@{}ll@{}}
    \toprule
    \textbf{Construction} & \textbf{Example Sentence} \\
    \midrule
    \textbf{LA} (Let Alone)         & I can't knit a scarf, \textit{let alone sew a quilt}. \\
    \textbf{CC} (Comparative Correlative) & The faster you run, the sooner you tire. \\
    \textbf{CWT} (Caused Motion with Theme) & She filled the bucket with sand. \\
    \textbf{CON} (Conative)         & The boxer \textit{punched at} the heavy bag. \\
    \textbf{WAY} (Way Construction) & She danced her way to fame. \\
    \textbf{IM} (Intransitive Motion) & The children ran into the park. \\
    \textbf{CM} (Caused Motion)     & They rolled the log into the river. \\
    \textbf{RES} (Resultative)      & He hammered the metal flat. \\
    \bottomrule
    \end{tabular}
    \caption{Eight challenge constructions ordered from most lexically substantive (top) to most schematic (bottom). Each example instantiates the construction in context.}
    \label{tab:construction-table}
    \end{table*}

\section{ConTest-NLI Templates}
\label{appendix:templates}

\begin{description}
  \item[Causative-With]
    \begin{itemize}
      \item \textbf{Prompt:} Describe a situation where something causes a place or thing to have a new feature or quality.\\
            \textbf{Example:} The party filled the room with laughter and music.
      \item \textbf{Prompt:} Write about an action that makes an object filled or loaded with something else.\\
            \textbf{Example:} She packed the suitcase with clothes for the trip.
      \item \textbf{Prompt:} Imagine someone causing a change by adding something to a space. Describe it.\\
            \textbf{Example:} They stocked the pantry with canned goods before the storm.
    \end{itemize}

  \item[Caused-Motion]
    \begin{itemize}
      \item \textbf{Prompt:} Write about someone making something move to a new place.\\
            \textbf{Example:} He pushed the broken car into the garage.
      \item \textbf{Prompt:} Describe an action that results in an object being relocated somewhere.\\
            \textbf{Example:} She threw the ball across the yard.
      \item \textbf{Prompt:} Tell a short story where an action causes an object to end up somewhere else.\\
            \textbf{Example:} The wind carried the leaves onto the porch.
    \end{itemize}

  \item[Comparative-Correlative]  
    \begin{itemize}
      \item \textbf{Prompt:} Describe a situation where two things change together — as one increases or decreases, so does the other.\\
            \textbf{Example:} The more he practiced, the better he became at playing the piano.
      \item \textbf{Prompt:} Write a sentence showing how more or less of one thing affects another thing.\\
            \textbf{Example:} The less she slept, the grumpier she got.
      \item \textbf{Prompt:} Imagine a cause-and-effect relationship where two actions or qualities are linked. Explain it.\\
            \textbf{Example:} The more it rained, the faster the river rose.
    \end{itemize}

  \item[Conative]
    \begin{itemize}
      \item \textbf{Prompt:} Write about someone trying to interact with an object but not necessarily succeeding fully.\\
            \textbf{Example:} He tugged at the door, but it wouldn't budge.
      \item \textbf{Prompt:} Describe an action where a person touches or tries to affect something without completely changing it.\\
            \textbf{Example:} She tapped at the microphone to check if it was working.
      \item \textbf{Prompt:} Imagine someone fiddling with or attempting to do something to an object — describe it.\\
            \textbf{Example:} He poked at the firewood, trying to get the flames to grow.
    \end{itemize}

  \item[Intransitive Motion]
    \begin{itemize}
      \item \textbf{Prompt:} Describe a person, animal, or thing moving from one place to another.\\
            \textbf{Example:} The cat wandered into the kitchen.
      \item \textbf{Prompt:} Write about a movement where the focus is on someone or something changing location.\\
            \textbf{Example:} The children raced down the hill.
      \item \textbf{Prompt:} Tell a short story about a journey or movement from one place to another.\\
            \textbf{Example:} The balloon drifted across the blue sky.
    \end{itemize}

  \item[Let-Alone]
    \begin{itemize}
      \item \textbf{Prompt:} Describe a situation where something is already hard or unlikely — and an even harder thing is even less likely.\\
            \textbf{Example:} He couldn't finish a page of his homework, let alone the entire assignment.
      \item \textbf{Prompt:} Write about two related actions or qualities, where the second is even more extreme than the first.\\
            \textbf{Example:} I can barely manage to jog a mile, let alone run a marathon.
      \item \textbf{Prompt:} Imagine someone struggling with one task — and an even harder task is even more impossible. Describe it.\\
            \textbf{Example:} She had trouble cooking pasta, let alone baking a soufflé.
    \end{itemize}

  \item[Resultative]
    \begin{itemize}
      \item \textbf{Prompt:} Describe an action that causes something to change its state or condition.\\
            \textbf{Example:} He wiped the counter clean.
      \item \textbf{Prompt:} Write about an event where someone does something that makes an object end up different than before.\\
            \textbf{Example:} She hammered the metal flat.
      \item \textbf{Prompt:} Tell a story where an object transforms because of someone's actions.\\
            \textbf{Example:} They painted the walls bright yellow.
    \end{itemize}

  \item[Way-Manner]
    \begin{itemize}
      \item \textbf{Prompt:} Describe someone making progress by doing an action repeatedly or in a special way.\\
            \textbf{Example:} He elbowed his way through the crowded hallway.
      \item \textbf{Prompt:} Write about someone moving through space by performing an activity along the way.\\
            \textbf{Example:} She laughed her way down the mountain trail.
      \item \textbf{Prompt:} Imagine someone reaching a destination while doing something unusual — describe it.\\
            \textbf{Example:} They danced their way to the front of the stage.
    \end{itemize}
\end{description}

\section{Hyperparameters}
\label{appendix:hyperparameters}
Hyperparameters and justifications are given in Table~\ref{tab:generation_parameters}.

\begin{table*}[ht]
\centering
\begin{tabular}{|l|l|p{6.5cm}|}
\hline
\textbf{Parameter} & \textbf{Value} & \textbf{Justification} \\
\hline
\texttt{temperature} & 0.8 & maximises lexical variety without destabilising syntax \\
\hline
max tokens & 500 & covers premise + three hypotheses with margin \\
\hline
rare-lemma seed list & 5 376 nouns/verbs/adjectives & reduces overlap with pre-training corpora \\
\hline
\end{tabular}
\caption{Generation parameters and their justification.}
\label{tab:generation_parameters}
\end{table*}

\section{Fine-Tuning Details}
\label{appendix:fine-tuning}

Given the size of our labeled fine-tuning data, full fine-tuning of all model parameters 
would be computationally expensive, prone to overfitting, and very inflexible for our 
experiments. Therefore, we employ LoRA;
This approach significantly mitigates the risk of overfitting on smaller, highly 
structured datasets like ours.

LoRA modules (rank r=16, scaling factor $\alpha$=32, dropout p=0.05) are specifically 
injected into the attention layers and multi-layer perceptron projections of layers 12 
through 20 of the Llama-3-8B-Instruct model.

Layers 12-20 in a 32-layer transformer model, such as Llama 3.1 8B, are roughly in the 
middle of the network. Prior research shows that middle and upper-middle layers often encode 
a mix of syntactic and semantic abstractions - ideal for adapting models to semantic tasks 
like NLI, especially for constructional generalization \citep{liu2019linguistic}.

We also note that training is conducted using mixed-precision, with weights in 
\textsc{bfloat16} and 
activations in \textsc{int8}, to further reduce memory footprint and improve training 
efficiency. Additional fine tuning hyperparameters are found in Table 
\ref{tab:ft-hyperparameters}.

\begin{table*}[!ht]
    \centering
    \begin{tabular}{ll}
        \toprule
        \multicolumn{2}{c}{\textbf{Optimizer and Hyperparameters}} \\
        \midrule
        Optimizer & AdamW \\
        $\beta_1$ & $0.9$ \\
        $\beta_2$ & $0.999$ \\
        $\epsilon$ & $1e-8$ \\
        \midrule
        \multicolumn{2}{c}{\textbf{Regularization and Early Stopping}} \\
        \midrule
        BookCorpus in minibatches & $33\%$ (for anti-forgetting) \\
        BookCorpus loss weight & $0.25$ \\
        Label smoothing (NLI heads) & $0.1$ \\
        Gradient clipping (max norm) & $1.0$ \\
        Weight decay (LoRA matrices) & $0.01$ \\
        \bottomrule
    \end{tabular}
    \caption{Hyperparameters from fine-tuning experiments}
    \label{tab:ft-hyperparameters}
\end{table*}

\section{Training Regimes}
\label{appendix:training-regimes}

We use three distinct training regimes designed to disentangle the effects of weight updates, classifier head architecture, and dataset characteristics.

\begin{description}
    \item[Shared-Head:] Updates the model with a single shared three-way NLI 
    head for all constructions. This is the canonical regime, testing if a unified 
    representation can be learned across all constructions.
    
    \item[Per-construction Heads:] Updates the model with 8 independent NLI heads 
    (one per construction). This explores whether separate, specialized classifier heads 
    better capture constructional nuances.
    
    \item[In-Context Few-Shot:] No weights are updated. Predictions are made via prompting 
    (8-shot). This baseline tests learning from examples in context, without training.
\end{description}

All fine-tuning regimes are run for a maximum of 5 epochs over the training data. 
By comparing performance across these 
regimes, we can draw more nuanced conclusions: for example, if the Shared-Head regime 
significantly outperforms In-Context Few-Shot, it suggests that explicit weight updates are 
beneficial. The Per-construction Heads condition offers insights into the potential modularity of 
learned constructional knowledge. This comprehensive experimental design ensures that claims 
about improved constructional understanding are robust and well-substantiated. 

\section{Full Error Analysis}
\label{appendix:error-analysis}

The examples extracted and displayed in table \ref{tab:errors} each illustrate a distinct type of model failure identified in our study. 

\begin{small}
\begin{table*}[ht!]
\centering
\begin{tabular}{p{2cm}p{4cm}p{4cm}p{1.8cm}p{2.1cm}}
\toprule
\textbf{Construction} & \textbf{Premise} & \textbf{Hypothesis} & \textbf{Gold Label} & \textbf{Model Label} \\
\midrule
Conative & The carpenter repeatedly hammered at the stubborn nail. & The carpenter gave up trying to fix the nail. & Contradiction & Neutral \\
Way-manner & The detective elbowed his way to the front of the crowded room. & The detective stayed at the back of the room. & Contradiction & Neutral \\
Caused-motion & The artist painted the mural into a vibrant masterpiece. & The artist worked on the mural for a week. & Neutral & Entailment \\
Let-alone & He barely managed to tie his shoelaces, let alone complete the marathon. & He found tying his shoelaces easy. & Contradiction & Neutral \\
Intransitive-motion & Without a destination, the traveler wandered through the forest. & The traveler had a clear destination in mind. & Contradiction & Entailment \\
Resultative & A few strikes were enough: the blacksmith hammered the iron flat. & The iron became flat. & Entailment & Neutral \\
\bottomrule
\end{tabular}
\caption{Examples of failed NLI cases from the ConTest-NLI training set.}
\label{tab:errors}
\end{table*}
\end{small}

In the Conative and Way-manner cases, the model recognised the action but failed to apply constructional entailments --- ongoing effort should contradict ``gave up'', and the Way construction implies location change. The Caused-motion and Resultative examples show that the model often conflates transformation events with generic processes, ignoring the causative semantics that the construction encodes. The Let-alone example reveals a missed scalar inference, treating ``barely managed'' as isolated from the second clause. Finally, the Intransitive-motion case highlights a negation cue failure, where ``without a destination'' was incorrectly aligned with a positive statement due to lexical overlap. Across constructions, these failures demonstrate that improvements from fine-tuning largely reflect memorisation of surface patterns rather than abstraction of form–meaning pairings.

\end{document}